\documentclass[letterpaper, 10 pt, conference]{ieeeconf}  
\IEEEoverridecommandlockouts                              \usepackage{caption}        
\usepackage{xcolor}
\long\def\invis#1{}
\usepackage [english]{babel}
\usepackage [autostyle, english = american]{csquotes}
\MakeOuterQuote{"}
\usepackage{booktabs}
\usepackage[colorlinks]{hyperref}
\usepackage{wrapfig}
\usepackage{amssymb}

\usepackage{subcaption}
\usepackage{amsmath}

\usepackage{bbm}
\usepackage{algpseudocode}
\usepackage{algorithm}
\usepackage{graphicx}
\usepackage{cite}
\usepackage{siunitx}
\usepackage{float}

\newcommand{\LAVA}{\textbf{LAVA}}
\renewcommand\paragraph{\@startsection{paragraph}{4}{\z@}%
            {-2.5ex\@plus -1ex \@minus -.25ex}%
            {1.25ex \@plus .25ex}%
            {\normalfont\normalsize\bfseries}}

\usepackage{xcolor}

\title{\LARGE \bf
\textsc LAVA: Long-horizon Visual Action based Food Acquisition
}

\author{Amisha Bhaskar, Rui Liu, Vishnu D. Sharma, Guangyao Shi, Pratap Tokekar   % <-this % stops a space
\thanks{All authors are from the University of Maryland, College Park, MD 20742 USA. \tt\small \{amishab, ruiliu, vishnuds, gyshi, tokekar\}@umd.edu}}

\begin{document}

\maketitle
\thispagestyle{empty}
\pagestyle{empty}

%%%%%%%%%%%%%%%%%%%%%%%%%%%%%%%%%%%%%%%%%%%%%%%%%%%%%%%%%%%%%%%%%%%%%%%%%%%%%%%%

\begin{abstract}

Robotic Assisted Feeding (RAF) addresses the fundamental need for individuals with mobility impairments to regain autonomy in feeding themselves. The goal of RAF is to use a robot arm to acquire and transfer food to individuals from the table. Existing RAF methods primarily focus on solid foods, leaving a gap in manipulation strategies for semi-solid and deformable foods. This study introduces Long-horizon Visual Action (\LAVA{}) based food acquisition of liquid, semisolid, and deformable foods. Long-horizon refers to the goal of "clearing the bowl" by sequentially acquiring the food from the bowl. \LAVA{} employs a hierarchical policy for long-horizon food acquisition tasks. The framework uses high-level policy to determine primitives by leveraging ScoopNet. At the mid-level, \LAVA{} finds parameters for primitives using vision. To carry out sequential plans in the real world, \LAVA{} delegates action execution which is driven by Low-level policy that uses parameters received from mid-level policy and behavior cloning ensuring precise trajectory execution.

We validate our approach on complex real-world acquisition trials involving granular, liquid, semisolid, and deformable food types along with fruit chunks and soup acquisition. Across 46 bowls, \LAVA{} acquires much more efficiently than baselines with a success rate of $89\pm4\%$, and generalizes across realistic plate variations such as different positions, varieties, and amount of food in the bowl. Code, datasets, videos, and supplementary materials can be found on our \href{https://raaslab.org/projects/RoboSpoon/}{website}.

\end{abstract}

\section{INTRODUCTION}
\label{section:introduction}
%%%%%%%%%%%%%%%%%%%%%%%%%%%%%%%%%%%%%%%%%%%%%%%%%%%%%%%%%%%%%%%
For individuals with limited mobility or disabilities, the act of feeding themselves can pose a significant challenge. This challenge has motivated the development of Robotic Assisted Feeding (RAF) \cite{brose2010role} aiming to restore independence and enhance the quality of life for those affected, while also alleviating the caregiver burden. A key component of such an assistive feeding system is bite acquisition, i.e., the act of a robotic arm picking up morsels of food from a plate to transfer the food to a person’s mouth. 

Navigating the diverse array of foods—from granular cereals to semi-solid food such as yogurt and deformable food items such as tofu---without breakage or deformation presents significant challenges for RAF \cite{grannen2022learning,sundaresan2023learning}. Additionally, the dynamic positioning of food chunks within a fluid medium complicates the prediction of their exact location at the time of scooping, requiring sophisticated sensing and real-time adaptation capabilities. This underscores the need for RAF systems to exhibit not only dexterity but also an advanced understanding of the dynamic nature of various food types and operate over a long horizon to clear the bowl.

Traditional RAF methodologies have depended on hard-coded adaptation strategies and low-level vision-parametrized primitives for food manipulation, employing distinct tools and primitives for specific tasks such as skewering \cite{gallenberger2019transfer,feng2019robot,8624330,sundaresan2022learning}, bite transfer \cite{belkhale2022balancing,gallenberger2019transfer,jenamani2024feel}, scooping \cite{grannen2022learning} and even end-to-end system \cite{gordon2024adaptable, Meet_obi}. 

%%%%%%%%%%%%%%%%%%%%%%%%%%%%%%%%%%%%%%%%%%%%%%%%%%%%%%%%%%%%%%%
\begin{figure}[t]
    \centering
    {\includegraphics[width=1\columnwidth]{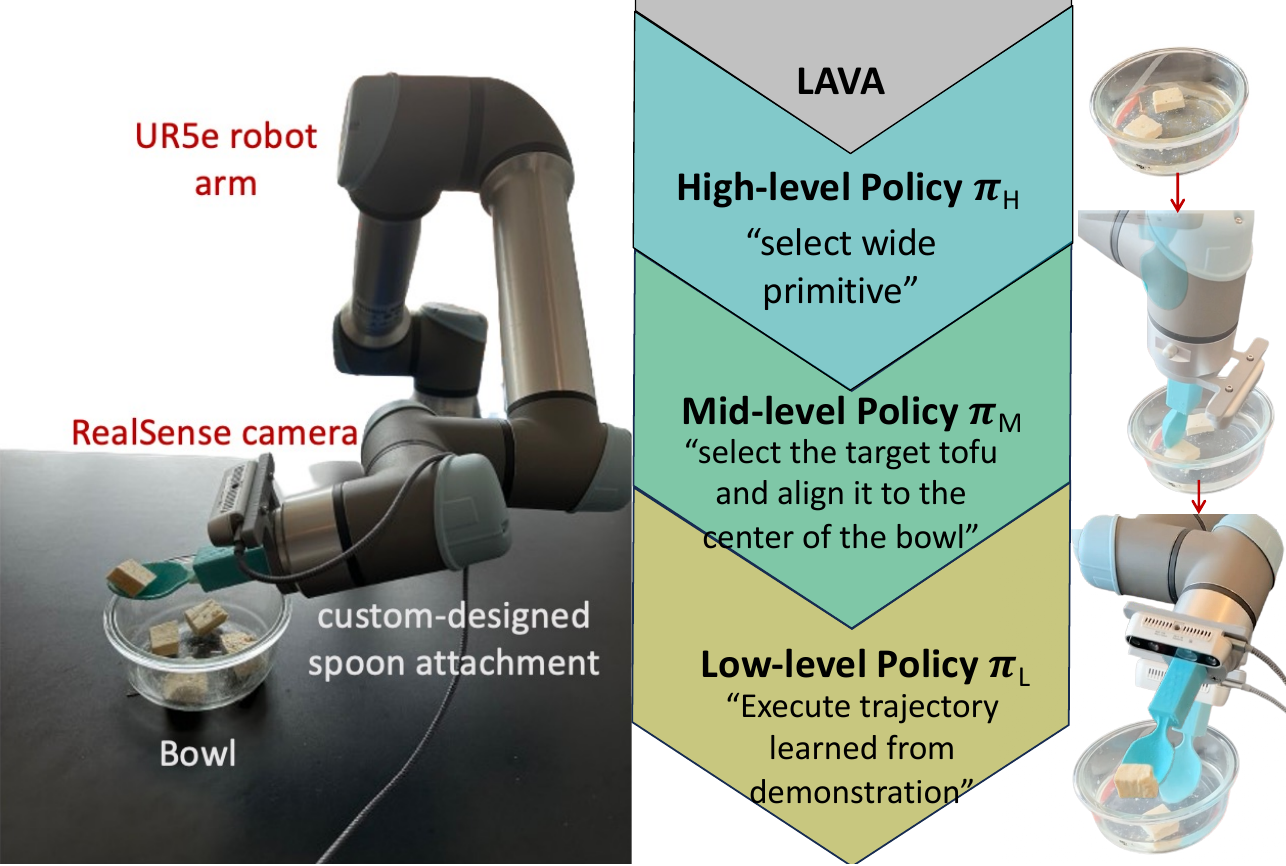}}
    % \vspace{-1em}
    \caption{System setup for LAVA alongside an illustrative description of the proposed framework with snapshots of task execution.}
    % \vspace{-1.5em}
    \label{fig:experimental_setup}
\end{figure}

%%%%%%%%%%%%%%%%%%%%%%%%%%%%%%%%%%%%%%%%%%%%%%%%%%%%%%%%%%%%%%%
%%%%%%%%%%%%%%%%%%%%%%%%%%%%%%%%%%%%%%%%%%%%%%%%%%%%%%%%%%%%%%%
\begin{figure*}[htbp]
    \centering
    \includegraphics[width=\textwidth]{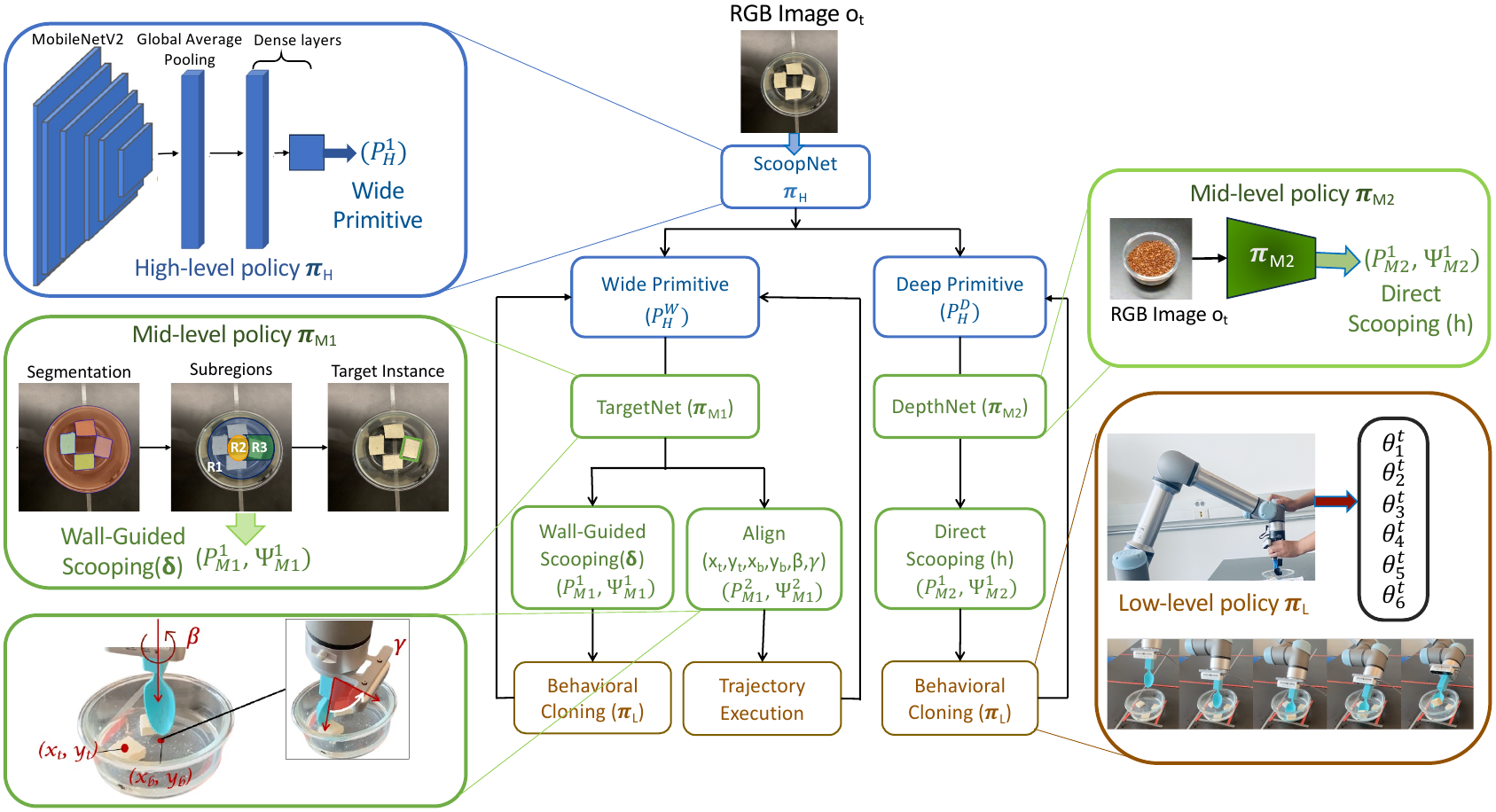}
    \caption{\textbf{\LAVA{}}: System Architecture of LAVA wich employs a high level policy(blue) $\pi_H$ to select amongst discrete high level primitives $P_{H}^{k}$, such as wide primitive and Deep primitive, which then further gets refined by mid-level policy (green) $\pi_M$ to select amongst mid-level primitives$P_{M}^{k}$, low-level vision parametrized policy $\pi_L$ (brown) executes trajectory learned from Behavioral cloning for long-horizon dextrous food acquisition.}
    % \vspace{-1.5em}
    \label{fig:setup}
\end{figure*}

%%%%%%%%%%%%%%%%%%%%%%%%%%%%%%%%%%%%%%%%%%%%%%%%%%%%%%%%%%%%%%%
This approach, while effective for singular, isolated actions, falls short in replicating the complex, sequential behaviors exhibited by humans during feeding. Humans adeptly combine various actions, such as scooping both solid chunks and liquid from a bowl in a single motion or rearranging food items for easier acquisition, demonstrating a nuanced understanding and strategy that spans the entire meal. This limitation underscores a gap in RAF technology and highlights the need for an advanced understanding and replication of human-like, long-horizon feeding strategies capable of managing both the rigidity of solid foods and the complexity of deformable items.

Recent advancements in skill-based reinforcement learning (RL) offer promising methodologies for modeling these complex, long-horizon manipulation sequences in a hierarchical manner. This entails first learning a high-level policy for composing
skills \cite{lin2022planning}, and then optionally inferring the parameters of low-level skills separately \cite{dalal2021accelerating,nasiriany2022augmenting}. Such approaches have shown potential, yet they face limitations when applied to the food domain, which demands high-fidelity models for food deformation, visual recognition, and utensil interaction not fully captured in current simulations. While VAPORS \cite{sundaresan2022learning} demonstrates effective long-horizon planning for specific food items such as noodles, it relies heavily on simulation for learning plate dynamics and lacks adaptability and applicability in real-world scenarios for broader categories of food types.

Thus, we seek to find an appropriate layer of abstraction for feeding, which can leverage the benefits of (1) hierarchical planning for long-horizon manipulation; (2) vision-based primitives for fine-grained control; and (3) flexible approach that can dynamically adapt to the wide variety of challenges presented by different food types.

Recognizing these challenges, this work introduces \LAVA{} (Long-horizon Acquisition via Visual Action), as a hierarchical policy for sequential planning of food acquisition (see Figure \ref{fig:experimental_setup}). Our approach is decoupled into three level-policy:  a high-level policy that identifies primitives based on visual inputs; a mid-level policy to refine these primitives and parameterize the actions of the lower-level policy and a low-level policy to use those parameters to rearrange and acquire food items and sequentially clear the bowl. 

The key contributions of this paper are:
\begin{itemize}
    \item We present a comprehensive hierarchical policy framework for long-horizon, visual-action-based food acquisition that systematically divides the task of food acquisition into high-level decision-making, mid-level action refinement, and low-level execution. 
    \item  Our method showcases adaptability and robustness across a diverse range of food types, effectively clearing bowls. It addresses and surpasses previous limitations in adaptability to various food types and the challenge of completing meals.
    \item We introduce a dataset of food items, showcasing different volumes and spatial arrangements within the bowl.
    \item We evaluate the learned scooping policies through real-world deployment system with UR5e and end-effector coupled with spoon attachment and D435i RealSense camera on the wrist. 
\end{itemize}

The remaining of the paper is organized as follows. In Section \ref{section:related_work}, we review related work on robotic assisted feeding, imitation learning, and long-horizon planning. In Section \ref{section:problem_statement}, we present the problem statement. In Section \ref{section:proposed_approach}, we introduce the multi-level policy including high-level policy, mid-level policy, and low-level policy. In Section \ref{section:experiments}, we present and discuss the real robot experimental results and compare them with the baseline. Finally, Section \ref{section:conclusion}, concludes the paper with lessons learned and possible future work.

\section{RELATED WORK}
\label{section:related_work}

We build on prior works studying multisensory robot learning and long-horizon task planning both within and beyond the food domain. In this section, we will discuss related work in robot-assisted feeding, learning from demonstration, and more generally long-horizon planning and control.
%%%%%%%%%%%%%%%%%%%%%%%%%%%%%%%%%%%%%%%%%%%%%%%%%%%%%%%%%%%%%%%
\subsection{Robotic-Assisted Feeding}
Robotic Assisted Feeding (RAF) can be split into two stages: bite acquisition and bite transfer. Previous work in RAF focused on bite acquisition and transfer with the aid of robotic arms and specialized tools such as spoons and forks \cite{belkhale2022balancing,bhattacharjee2019towards,gallenberger2019transfer,8624330,grannen2022learning}. The incorporation of computer vision has enabled these systems to adapt to various food types and user preferences, with models such as SPANet \cite{feng2019robot} demonstrating proficiency in mapping food images to actions. However, challenges remain in handling semi-solid and deformable foods, where generalizable strategies are scarce and bimanual scooping \cite{grannen2022learning} techniques have shown limited success. Market-available devices \cite{Meet_obi} offer mealtime assistance but are constrained by their reliance on teleoperation and the physical limitations of their design. In isolation, this does not capture
many long-horizon real-world feeding scenarios with multiple utensils and strategies.
While prior research has made strides in visual planning and manipulation for specific food items \cite{sundaresan2022learning}, a comprehensive approach that addresses the adaptability to a wide array of food types and real-world feeding scenarios is still needed.
%%%%%%%%%%%%%%%%%%%%%%%%%%%%%%%%%%%%%%%%%%%%%%%%%%%%%%%%%%%%%%%

%%%%%%%%%%%%%%%%%%%%%%%%%%%%%%%%%%%%%%%%%%%%%%%%%%%%%%%%%%%%%%%
\subsection{Learning from Demonstration}
Learning from Demonstration (LfD) is a methodology where robots learn new skills by observing expert demonstrations, which can be performed by humans or intelligent agents. This approach is particularly useful for tasks that are challenging to pre-program but can be easily demonstrated. LfD has been applied across various domains, including robotic assembly in manufacturing \cite{zhu2018robot}, path planning for complex tasks \cite{xie2020robot}, assistive technologies in rehabilitation \cite{lauretti2017learning}, and intricate picking and placing tasks \cite{rahmatizadeh2018vision}. The technique is divided into three main approaches: kinesthetic teaching \cite{akgun2012trajectories}, where a human physically guides the robot; teleoperation \cite{si2021review}, where the robot is remotely controlled; and passive observation \cite{vogt2017system}, where the robot learns by watching. Our research primarily utilizes kinesthetic teaching to instruct a UR5e robot arm in scooping tasks. Within LfD's learning objectives, our focus is on developing policies for handling semi-solid and deformable food items, and optimizing their scooping trajectories.

% —learning a policy, learning a reward or cost function, and learning a plan—

%%%%%%%%%%%%%%%%%%%%%%%%%%%%%%%%%%%%%%%%%%%%%%%%%%%%%%%%%%%%%%%

%%%%%%%%%%%%%%%%%%%%%%%%%%%%%%%%%%%%%%%%%%%%%%%%%%%%%%%%%%%%%%%
\subsection{Long-Horizon Planning and Control}
Recent works in long-horizon manipulation frameworks have explored separating high-level strategic decision-making from detailed motion planning. Traditional task-and-motion planning approaches rely on extensive domain knowledge and fixed task sequences \cite{srivastava2014combined,garrett2021integrated,chitnis2016guided}, but falter due to the unpredictable dynamics of food on a plate and the complexity of state estimation. Model-based planning has shown promise in tasks such as dough manipulation by using environment dynamics learned from visual inputs to plan action sequences \cite{shi2023robocraft,lin2022planning}. However, these methods struggle with the high-dimensional action spaces typical in food acquisition. Hierarchical reinforcement Learning offers a solution by dividing decision-making into high-level strategic planning and execution by discrete, parameterized low-level primitives \cite{pateria2021hierarchical}. While promising in simulation for tasks such as tabletop manipulation \cite{nasiriany2022augmenting,dalal2021accelerating}, these methods have limitations in real-world application and handling the diverse and complex manipulation tasks required for effective feeding. Our work aims to address these gaps by focusing on the adaptation to real-world scenarios and the development of specialized primitives for a wide variety of food items, challenging the scalability of current approaches and introducing the necessity for innovative solutions in robotic feeding.
\section{Problem Statement}
 \label{section:problem_statement}
In this work, we tackle the challenge of sequential bite acquisition to maximize the success rate and efficiency of long-horizon food acquisition to ensure efficient bowl clearance. The focus is on a variety of food types, ranging from granular items such as cereals to semi-solid foods such as yogurt, and deformable substances such as tofu, all within a bowl fixed in position and assumed to be scoopable with a spoon.

We assume access to bowl image observations $o \in \mathbf{R_+}^{W \times H \times C} = \mathcal{O}$ of unknown bowl states S. Here, $W$, $H$, and $C$ denote the image dimensions. The image is sourced from a camera attached to the wrist of the robotic arm with a custom spoon attachment as an end-effector. We have access to expert demonstration data for robot proprioceptive information (joint positions). Our goal is to learn a policy $\pi(\phi_t|o_t)$ that takes RGB images as input $(o_t)$ and returns output as joint angles $\theta_t$ of the arm for efficient long-horizon food acquisition. In this context, long-horizon refers to a series of sequential actions aimed at complete bowl clearance.

\section{Proposed Approach}
\label{section:proposed_approach}

%%%%%%%%%%%%%%%%%%%%%%%%%%%%%%%%%%%%%%%%%%%%%%%%%%%%%%%%%%%%%%%%%%%%%%%%%%%%%%%%%%%%%%

We formalize the long-horizon food acquisition setting as a hierarchical policy $\pi$. To do so we decouple $\pi$ into separate high, mid, and low-level sub-policies. We assume access to $K$ discrete manipulation primitives $P_{H}^{k}$, $k \in {1,..., K}$, and learn a high-level policy $\pi_{H}$ which selects amongst these primitives based on visual input $o_t$. The mid-level policy $\pi_M$ further refines this selection, parameterizing the low-level policy $\pi_L$ based on both the chosen primitive and additional visual inputs. 

This low-level policy then executes a sequence of actions $\theta_{t}^{k}$, aimed at achieving precise food acquisition. This hierarchical arrangement is formalized as follows:

\begin{itemize}

    \item \textbf{High-level policy:} $\pi_H(P_{H}^{k}|o_t)$ focuses on selecting the manipulation primitive suitable for the current visual scene.

    \item \textbf{Mid-level policy:} $\pi_M(P_{M}^{k}, \psi_{M}^{k}|o_t, P_{H}^{k})$ refines this choice by parameterizing actions tailored to the specific food item's characteristics.

    \item \textbf{Low-level policy:} $\pi_L(\theta_{t}^{k}|P_{M}^{k},\psi_{M}^{K})$ executes the action sequence, utilizing parameters and primitives from the mid-level policy.

\end{itemize}

We consider low-level actions $\theta_t$, parameterized by the position of the tip of a spoon $(x, y)$ and spoon roll and pitch $(\gamma, \beta)$ in the wrist frame of reference. As shown in Figure \ref{fig:setup} detailing the LAVA setup, we describe each module in LAVA in further detail.

%%%%%%%%%%%%%%%%%%%%%%%%%%%%%%%%%%%%%%%%%%%%%%%%%%%%%%%%%%%%%%%%%%%%%%%%%%%%%%%%%%%%%%

%%%%%%%%%%%%%%%%%%%%%%%%%%%%%%%%%%%%%%%%%%%%%%%%%%%%%%%%%%%%%%%%%%%%%%%%%%%%%%%%%%%%%%
\subsection{High-level Policy}
At the highest level of our hierarchical model, the high-level policy $\pi_H(P_{H}^{k}|o_t)$ uses visual cues to select the most suitable scooping primitive---Wide Primitive ($P_{H}^{W}$) and Deep Primitive ($P_{H}^{D}$), based on the food type present.

%%%%%%%%%%%%%%%%%%%%%%%%%%%%%%%%%%%%%%%%%%%%%%%%%%%%%%%%%%%%%%%%%%%%%%%%%%%%%%%%%%%%%%
\subsubsection{Wide Primitive ($P_{H}^{W}$)}
Wide Primitive, is a strategy developed for handling foods that lack cohesion or are deformable, such as tofu or certain types of jelly. This method involves using the bowl's wall as a guide and support mechanism for the scooping action. By gently pressing the food against the wall of the bowl, it creates a pseudo-cohesive mass that can be scooped more easily. This technique is especially useful for foods that tend to scatter or break apart, as the wall provides the necessary containment to gather and scoop the food effectively. Instance scooping requires sophisticated control over the spoon's movement, including adjusting the angle applied against the food and the bowl wall, to achieve the desired outcome without compromising the integrity of the food or missing the target. It requires identifying the target instance to not collide with other instances or break them in the process. The wide primitive is implemented with the other two mid-level primitives align and wall-guided scooping described in Section~\ref{subsubsection:TN}
%%%%%%%%%%%%%%%%%%%%%%%%%%%%%%%%%%%%%%%%%%%%%%%%%%%%%%%%%%%%%%%%%%%%%%%%%%%%%%%%%%%%%%

\begin{figure}[ht]
    \centering
    \includegraphics[width=0.5\textwidth]{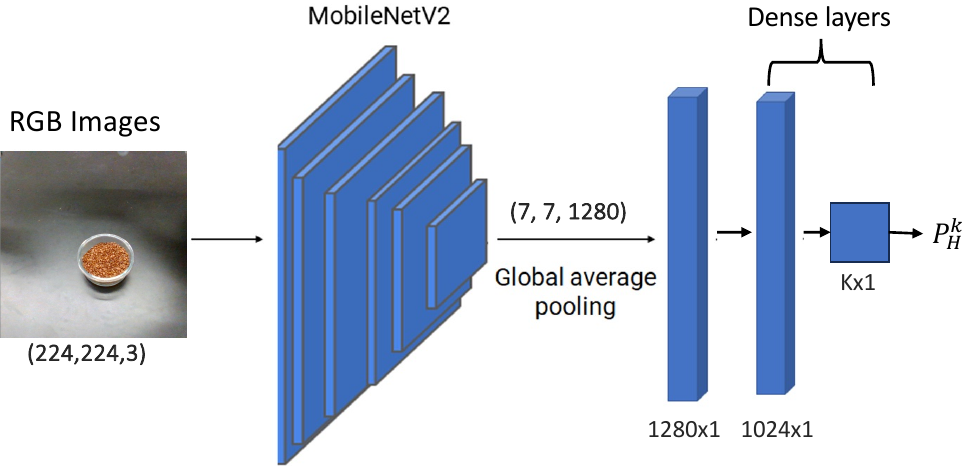}
    % \vspace{-1em}
    \caption{ScoopNet outputs the softmax probabilities over the high-level primitive depending on the type of food items present in the image.}
    % \vspace{-1.5em}
    \label{fig:scoopNet}
\end{figure}
%%%%%%%%%%%%%%%%%%%%%%%%%%%%%%%%%%%%%%%%%%%%%%%%%%%%%%%%%%%%%%%%%%%%%%%%%%%%%%%%%%%%%%
\subsubsection{Deep Primitive ($P_{H}^{D}$)}
Deep Primitive, on the other hand, is a straightforward approach designed for foods that possess enough cohesion to be picked up directly by a spoon without requiring additional support or manipulation. This method is particularly effective for liquid and semi-solid foods such as yogurt or porridge, where the food's natural consistency allows it to adhere to the spoon when scooped directly from the top or side. The key to successful direct scooping lies in the precise control of the spoon's trajectory and depth of penetration into the food, ensuring that a sufficient quantity is acquired without disturbing the remaining contents of the bowl excessively. The deep primitive is implemented with the direct scooping mid-level primitive described in Section~\ref{subsubsection:ds}
%%%%%%%%%%%%%%%%%%%%%%%%%%%%%%%%%%%%%%%%%%%%%%%%%%%%%%%%%%%%%%%%%%%%%%%%%%%%%%%%%%%%%%
%%%%%%%%%%%%%%%%%%%%%%%%%%%%%%%%%%%%%%%%%%%%%%%%%%%%%%%%%%%%%%%%%%%%%%%%%%%%%%%%%%%%%%
\subsubsection{ScoopNet ($\pi_H$)}
\label{subsection:ScoopNet}
ScoopNet is a network designed to select between the two high level primitives based on the type of food, utilizing the MobileNetV2 architecture \cite{Sandler_2018_CVPR} as the base. We train on a dataset of 5316 images from a custom collection and additional sources, targeting binary classification of high-level primitives $P_{H}^{1},..., P_{H}^{k}$. We used data augmentation (including rotations, zooms, and flips) to increase robustness against food image variations. Our dataset is available online.

%%%%%%%%%%%%%%%%%%%%%%%%%%%%%%%%%%%%%%%%%%%%%%%%%%%%%%%%%%%%%%%%%%%%%%%%%%%%%%%%%%%%%%

%%%%%%%%%%%%%%%%%%%%%%%%%%%%%%%%%%%%%%%%%%%%%%%%%%%%%%%%%%%%%%%%%%%%%%%%%%%%%%%%%%%%%%
The network is initially trained on the ImageNet dataset, with a customized final layer for specific task adaptation. This configuration, along with a Global Average Pooling layer and two dense layers ending in a sigmoid activation, uses the Adam optimizer and binary cross-entropy loss for accurate classification. The detailed architecture, ScoopNet, is depicted in Figure \ref{fig:scoopNet}. The output of ScoopNet is softmax probabilities over high-level primitives.

%%%%%%%%%%%%%%%%%%%%%%%%%%%%%%%%%%%%%%%%%%%%%%%%%%%%%%%%%%%%%%%%%%%%%%%%%%%%%%%%%%%%%%

%%%%%%%%%%%%%%%%%%%%%%%%%%%%%%%%%%%%%%%%%%%%%%%%%%%%%%%%%%%%%%%%%%%%%%%%%%%%%%%%%%%%%%

%%%%%%%%%%%%%%%%%%%%%%%%%%%%%%%%%%%%%%%%%%%%%%%%%%%%%%%%%%%%%%%%%%%%%%%%%%%%%%%%%%%%%%
\subsection{Mid-level Policy}
The Mid-level Policy $\pi_M(P_{M}^{k}, \psi_{M}^{K}|o_t, P_{H}^{k})$ serves as the intermediary layer that refines and parameterizes the chosen primitive for execution. This refinement is crucial for bridging the gap between high-level strategy selection and low-level action execution.
%%%%%%%%%%%%%%%%%%%%%%%%%%%%%%%%%%%%%%%%%%%%%%%%%%%%%%%%%%%%%%%%%%%%%%%%%%%%%%%%%%%%%%

%%%%%%%%%%%%%%%%%%%%%%%%%%%%%%%%%%%%%%%%%%%%%%%%%%%%%%%%%%%%%%%%%%%%%%%%%%%%%%%%%%%%%%
\subsubsection{TargetNet $(\pi_{M1})$ for Wide Primitive}
\label{subsubsection:TN}
\begin{figure}[ht]
    \centering
    \includegraphics[width=0.5\textwidth]{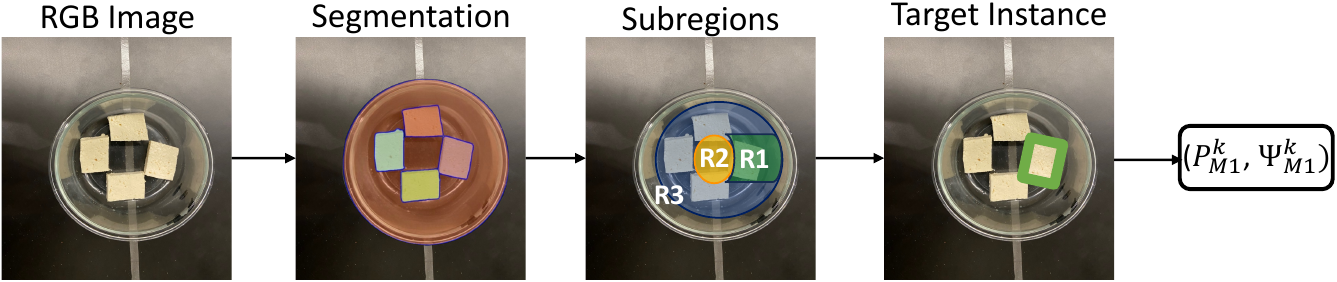}
    \caption{TargetNet finds the next ``target'' item for the wide high-level primitive and the mid-level primitive that decides whether to scoop the target item or to align it first.}
    % \captionsetup{belowskip=-10pt}
    \label{fig:targetNet}
    % \vspace{-1.5em}
\end{figure}

We have designed TargetNet, shown in Figure \ref{fig:targetNet}, that uses Mask R-CNN, tailored for the task of identifying and segmenting target food items such as tofu in a bowl, crucial for executing wide primitives. This model precisely segments food items, enabling the selection of appropriate mid-level primitives: wall-guided scooping and center align (described later in this section).

We use a custom dataset annotated for bowl, tofu, and target scooping areas, TargetNet employs transfer learning to accurately segment food items against diverse backgrounds, increasing its generalizability. In Section \ref{subsubsection:Zero-shot Generalization}, we report zero-shot generalization results for other types of food items. The model's training includes a COCOEvaluator to ensure segmentation accuracy meets COCO dataset \cite{cocodataset} standards.

%%%%%%%%%%%%%%%%%%%%%%%%%%%%%%%%%%%%%%%%%%%%%%%%%%%%%%%%%%%%%%%%%%%%%%%%%%%%%%%%%%%%%%
%%%%%%%%%%%%%%%%%%%%%%%%%%%%%%%%%%%%%%%%%%%%%%%%%%%%%%%%%%%%%%%%%%%%%%%%%%%%%%%%%%%%%%
Post-training, TargetNet creates a binary mask for pixels that are "occupied" by instances of food items. We divide the surrounding region of interest into sub-regions. If a sub-region intersects the bowl boundary, it is considered to be “occupied.” Otherwise, it is “unoccupied.” A food item is classified as “R1” if it is rightmost and closest to the wall, “R2” if the food item is at the center of the bowl, and “R3” otherwise. The subsequent visualization and centroid calculation steps of detected instances help with determining its location in subregions of the bowl and its location with respect to the center of the bowl, selecting between mid-level primitives \textemdash Wall-guided Scooping$(P_{M1}^{1})$ or Align$(P_{M1}^{2})$ and predicting parameters for low-level policy.
%%%%%%%%%%%%%%%%%%%%%%%%%%%%%%%%%%%%%%%%%%%%%%%%%%%%%%%%%%%%%%%%%%%%%%%%%%%%%%%%%%%%%%

%%%%%%%%%%%%%%%%%%%%%%%%%%%%%%%%%%%%%%%%%%%%%%%%%%%%%%%%%%%%%%%%%%%%%%%%%%%%%%%%%%%%%%
\textbf{Wall-guided Scooping $(P^1_{M1},\psi^1_{M1})$}
The Wall-guided Scooping strategy, parameterized by $\delta$—the centroid distance of the target instance from the bowl's center—adapts its approach based on the target's proximity to the bowl's wall and the sub-region. For food items in subregion R1, the strategy uses the wall's structural support for a scooping action. Conversely, items in central subregion R2 require a pre-scooping alignment, tactically moving the food towards the wall to simplify the scooping motion. 
%%%%%%%%%%%%%%%%%%%%%%%%%%%%%%%%%%%%%%%%%%%%%%%%%%%%%%%%%%%%%%%%%%%%%%%%%%%%%%%%%%%%%%

%%%%%%%%%%%%%%%%%%%%%%%%%%%%%%%%%%%%%%%%%%%%%%%%%%%%%%%%%%%%%%%%%%%%%%%%%%%%%%%%%%%%%%
\textbf{Align $(P^2_{M1},\psi^2_{M1})$}
The alignment step is essential for orienting the spoon to the target instance and guiding its movement toward the bowl's center. 
This procedure takes into consideration the centroid coordinates of the tofu ($x_t,y_t$) and the bowl's center ($x_b,y_b$) as well as the spoon's roll ($\gamma$) and pitch ($\beta$). Two key parameters are computed:

\begin{itemize}

    \item Spoon Orientation Angle: Calculated as $\gamma =\arctan(\frac{y_b - y_t}{x_b-x_t})$, this angle determines the necessary rotation of the spoon to align with the target instance, ensuring the spoon is positioned for optimal interaction and is untilted for planar push ($\beta= 0 {^\circ}$).

    \item Instance Push distance: Determined by ($x_t,y_t,x_b,y_b$), the instance is pushed from its current position towards the bowl's center, optimizing the positioning for the scooping action.

\end{itemize}
%%%%%%%%%%%%%%%%%%%%%%%%%%%%%%%%%%%%%%%%%%%%%%%%%%%%%%%%%%%%%%%%%%%%%%%%%%%%%%%%%%%%%%
%%%%%%%%%%%%%%%%%%%%%%%%%%%%%%%%%%%%%%%%%%%%%%%%%%%%%%%%%%%%%%%%%%%%%%%%%%%%%%%%%%%%%%
\subsubsection{DepthNet $(\pi_{M2})$ for Deep Primitive}
\begin{figure}[ht]
    \centering
    \includegraphics[width=\linewidth]{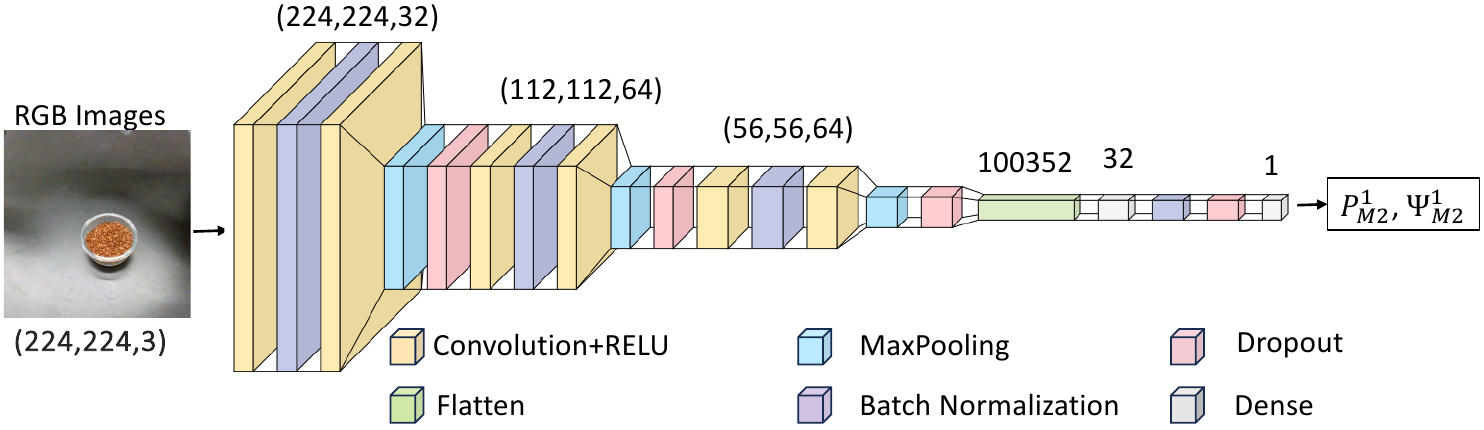}
    \caption{DepthNet detects the depth ($h$) of the food in the bowl.}
    % \captionsetup{belowskip=-10pt}
    \label{fig:depthNet}
    % \vspace{-1.0em}
\end{figure}

DepthNet is designed for depth detection of food in a bowl based on visual input $o_t$ and high-level primitive received from high-level policy. The architecture of DepthNet is outlined in Figure \ref{fig:depthNet}.

% to perform two tasks \textemdash height detection of food in a bowl and trajectory selection from behavioral cloning, based on visual input $o_t$ and high-level primitive received from high-level policy. The architecture of Depthnet is outlined in Figure \ref{fig:depthNet}.
%%%%%%%%%%%%%%%%%%%%%%%%%%%%%%%%%%%%%%%%%%%%%%%%%%%%%%%%%%%%%%%%%%%%%%%%%%%%%%%%%%%%%%
%%%%%%%%%%%%%%%%%%%%%%%%%%%%%%%%%%%%%%%%%%%%%%%%%%%%%%%%%%%%%%%%%%%%%%%%%%%%%%%%%%%%%%
DepthNet utilizes a vision-based approach to estimate the volume of food in a bowl, employing a Sequential model with convolutional layers of 32, 64, and 128 filters for feature extraction. These layers are augmented with batch normalization for improved training stability and dropout layers at rates of 0.25 and 0.5 to prevent overfitting. MaxPooling layers help reduce the dimensions of feature maps, increasing the model's efficiency. After convolutional processing, the model uses a flattened layer for data restructuring, followed by a dense layer with 32 neurons (using ‘relu' activation) for feature processing. The architecture culminates in a final dense layer with a single neuron (using ‘linear' activation) to predict the food's depth. DepthNet has been trained on a dataset of 1000 cereal images, categorized into three depth ranges in the bowl: 5.5 cm, 4 cm, and 2 cm, enabling precise depth estimation in varied food scenarios.

%%%%%%%%%%%%%%%%%%%%%%%%%%%%%%%%%%%%%%%%%%%%%%%%%%%%%%%%%%%%%%%%%%%%%%%%%%%%%%%%%%%%%%
\textbf{Direct scooping ($P^1_{M2},\psi^1_{M2}$)}\label{subsubsection:ds}
The direct scooping strategy employs a feedback mechanism centered on a predefined scooping axis, ($\beta= 0 {^\circ}$). The strategy utilizes the trained model on a dataset of correct trajectories taken by an expert human to scoop food from the bowl where the input is the position of the robotic arm relative to the bowl along with the estimated depth $(h)$ of the food received from DepthNet and the output is the adjusted trajectory from behavioral cloning based on inputs. This real-time adjustment is critical for achieving precise interaction between the scoop and the food item, ensuring effective scooping without causing displacement or spillage and long-horizon acquisition as the level of food changes while sequential scooping. Furthermore, this strategy is enhanced by the implementation of trajectory selection from behavioral cloning.
%%%%%%%%%%%%%%%%%%%%%%%%%%%%%%%%%%%%%%%%%%%%%%%%%%%%%%%%%%%%%%%%%%%%%%%%%%%%%%%%%%%%%%

%%%%%%%%%%%%%%%%%%%%%%%%%%%%%%%%%%%%%%%%%%%%%%%%%%%%%%%%%%%%%%%%%%%%%%%%%%%%%%%%%%%%%%
% \vspace{-0.5 em}
\subsection{Low-level policy}
We use Behavioral Cloning ($\pi_L$) with kinesthetic teaching to adapt scooping actions for different food textures and consistencies, improving the robot's performance in assistive feeding at the lowest level.  Various food items, with their unique requirements for scooping techniques, necessitate the modeling of distinct optimal scooping trajectories, especially for semi-solid and deformable foods. This process includes collecting demonstration data on joint positions, velocities, and timestamps to approach the scooping task as a trajectory optimization problem within the robot arm's joint space. 

The objective is to minimize a cost function $J(\tau)$ over a trajectory $\tau$, represented as $J(\tau) = \int_0^T L(\mathbf{q}(t), \dot{\mathbf{q}}(t)) dt$, where $\mathbf{q}(t)$ and $\dot{\mathbf{q}}(t)$ denote the robot's joint positions and velocities at time $t$, respectively, and $L(\cdot)$ is an instantaneous cost function penalizing deviations from the optimal trajectory.
The Weiszfeld algorithm \cite{beck2015weiszfeld,weiszfeld2009point} is used for this optimization, finding a trajectory $\hat{x}$ that minimizes the sum of Euclidean distances to demonstrated trajectories. It iteratively refines $\hat{x}$ until the adjustment falls below a small threshold $\epsilon$.

The algorithm updates the estimate of $\hat{x}$ using $\hat{x}_{k+1} = \frac{\sum_{i=1}^{n} \frac{p_i}{|\hat{x}_k-p_i|2}}{\sum_{i=1}^{n} \frac{1}{|\hat{x}_k-p_i|_2}}$, iterating until the change in $\hat{x}$ between iterations is below a predefined threshold $\epsilon$. This method determines optimal trajectories for the robot arm's joints, enhancing the robot's scooping accuracy and effectiveness. 
\section{Experiments}
\label{section:experiments}
%%%%%%%%%%%%%%%%%%%%%%%%%%%%%%%%%%%%%%%%%%%%%%%%%%%%%%%%%%%%%%%%%%%%%%%%%%%%%%%%%%%%%%

%%%%%%%%%%%%%%%%%%%%%%%%%%%%%%%%%%%%%%%%%%%%%%%%%%%%%%%%%%%%%%%%%%%%%%%%%%%%%%%%%%%%%%
For the experiments, we begin by describing the experimental setup. Following this, we discuss the data collection procedure and the baseline for comparison. Subsequently, we present and analyze the experimental results.
%%%%%%%%%%%%%%%%%%%%%%%%%%%%%%%%%%%%%%%%%%%%%%%%%%%%%%%%%%%%%%%%%%%%%%%%%%%%%%%%%%%%%%

%%%%%%%%%%%%%%%%%%%%%%%%%%%%%%%%%%%%%%%%%%%%%%%%%%%%%%%%%%%%%%%%%%%%%%%%%%%%%%%%%%%%%%
\subsection{Experimental Setup}

 The setup comprises a UR5e robot arm, a custom spoon attachment, a RealSense camera, and a fixed-position bowl, depicted in Figure \ref{fig:experimental_setup}. The spoon is affixed to the arm, with a length measuring $10.0\ cm$. The RealSense camera is attached to the wrist of the arm.

 During experiments, we explore varied configurations across the amount, size, position, and depth of food as well as different food types including granular cereals, liquid water, and semi-solid yogurt in the bowl. Food position configurations encompass multiple numbers of tofu and fruit chunks placed in different instance positions across the bowl. The varied amount and food depth included cereals, water, yogurt, and jelly filled at different depth levels inside the bowl. Additionally, we conduct tests with tofu chunks inside soup as shown in Figure \ref{fig:zero-shot}. For each food type, and depth, we conduct 10 trials of long-horizon food scooping attempts and for each position configuration in case of multiple tofu and fruit chunks, we conduct 5 trials of long-horizon food scooping attempts.
%%%%%%%%%%%%%%%%%%%%%%%%%%%%%%%%%%%%%%%%%%%%%%%%%%%%%%%%%%%%%%%%%%%%%%%%%%%%%%%%%%%%%%

%%%%%%%%%%%%%%%%%%%%%%%%%%%%%%%%%%%%%%%%%%%%%%%%%%%%%%%%%%%%%%%

%%%%%%%%%%%%%%%%%%%%%%%%%%%%%%%%%%%%%%%%%%%%%%%%%%%%%%%%%%%%%%%
%%%%%%%%%%%%%%%%%%%%%%%%%%%%%%%%%%%%%%%%%%%%%%%%%%%%%%%%%%%%%%%

%%%%%%%%%%%%%%%%%%%%%%%%%%%%%%%%%%%%%%%%%%%%%%%%%%%%%%%%%%%%%%%
\begin{figure}[htbp]
    \centering
    \begin{subfigure}[b]{0.47\linewidth}
        \centering
        \includegraphics[width=\linewidth]{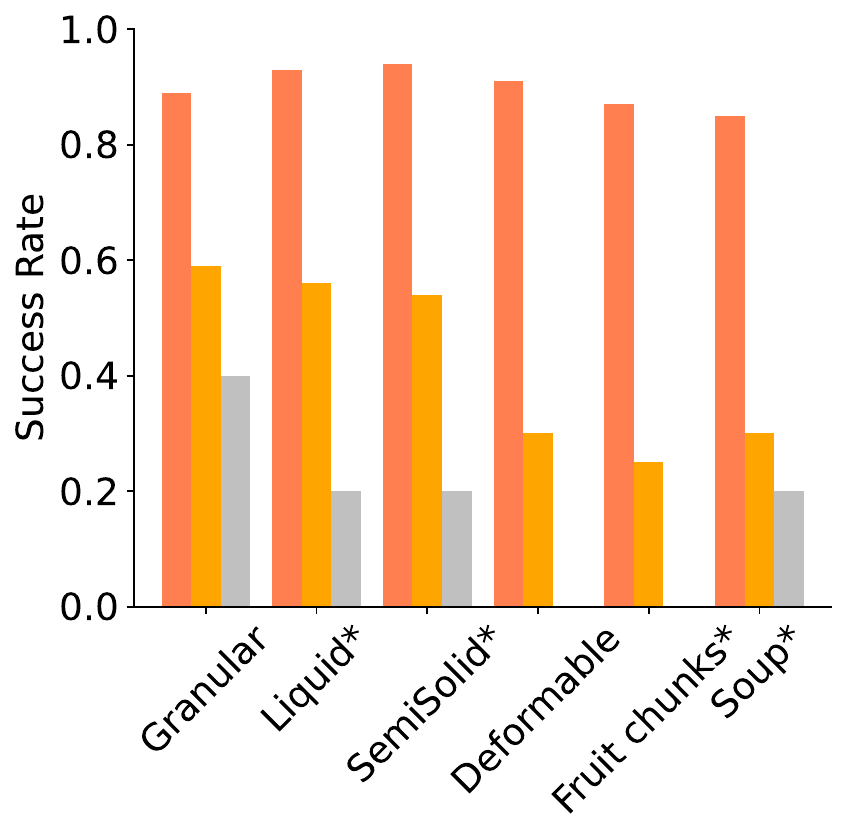}
        \caption{Overall Success Rate ($\uparrow$)}
        \label{fig:success_rate}
    \end{subfigure}
    \quad
    \begin{subfigure}[b]{0.47\linewidth}
        \centering
        \includegraphics[width=\linewidth]{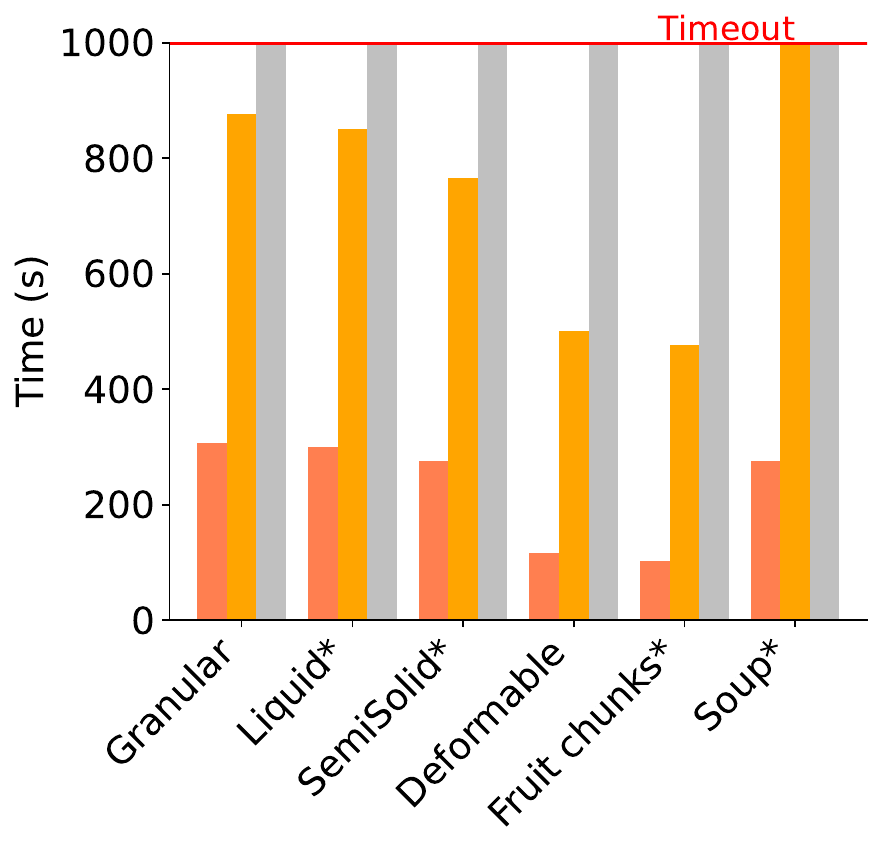}
        \caption{Total Time ($\downarrow$) }
        \label{fig:time}
    \end{subfigure}
    % \vspace{0.1em} % Adjust space between rows as needed

    \begin{subfigure}[b]{0.47\linewidth}
        \centering
        \includegraphics[width=\linewidth]{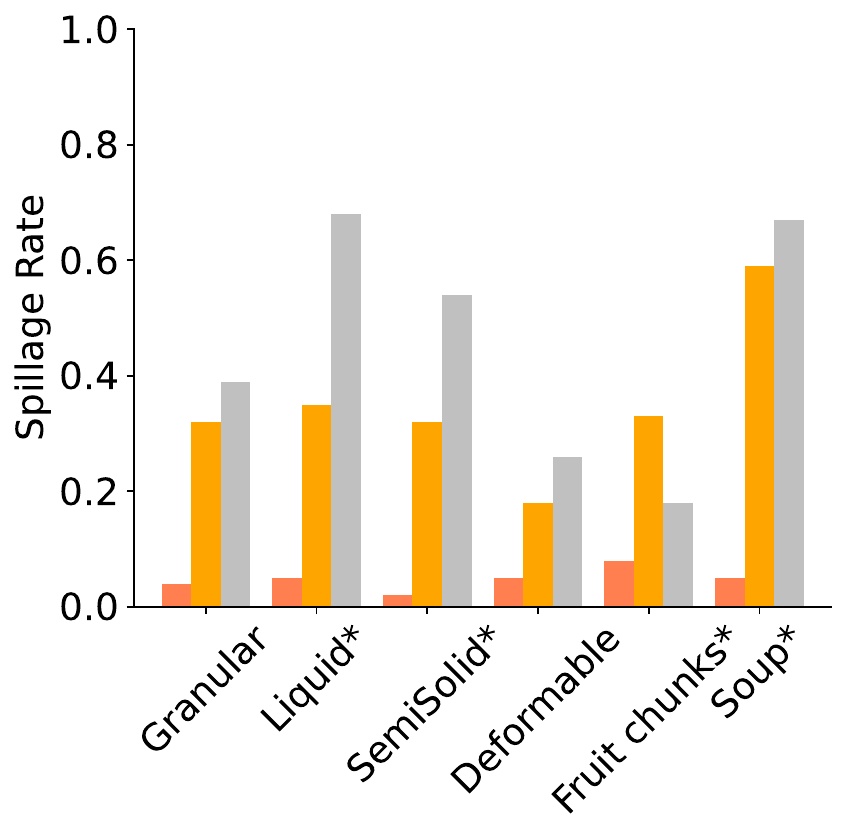}
        \caption{Spillage ($\downarrow$)}
        \label{fig:spillage}
    \end{subfigure}
    \quad
    \begin{subfigure}[b]{0.47\linewidth}
        \centering
        \includegraphics[width=\linewidth]{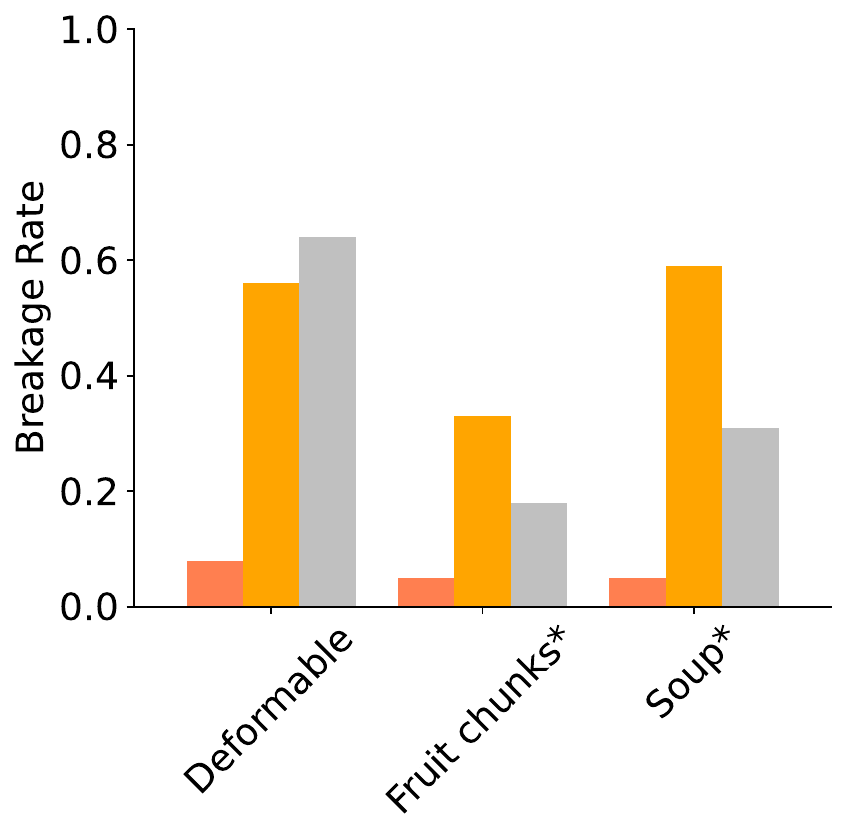}
        \caption{Breakage ($\downarrow$)}
        \label{fig:breakage}
    \end{subfigure}
    % \vspace{0.5em} % Adjust space before the legend

    % Include the common legend as an image
    \begin{subfigure}[b]{0.5\linewidth} % Use full text width for the legend
        \centering
        \includegraphics[width=\linewidth]{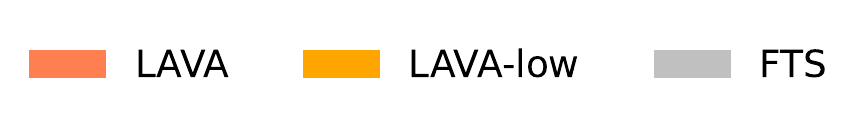} % Adjust the width as necessary
        % No caption for the legend, its understood to be common for all
    \end{subfigure}

    \caption{Breakdown of experimental performance comparison between \LAVA{}, LAVA-low, and Fixed Trajectory Scooping(FTS). $\ast$ represents zero-shot experiments.}
    % \vspace{-1em}
    \label{fig:subfigures_1}
\end{figure}

%%%%%%%%%%%%%%%%%%%%%%%%%%%%%%%%%%%%%%%%%%%%%%%%%%%%%%%%%%%%%%%%%%%%%%%%%%%%%%%%%%%%%%
\subsection{Data Collection}
\label{subsection:data_collection}

We collected data through kinesthetic teaching,  which encompassed two different types of trajectory--- wall-guided scooping and direct scooping, with twenty-five demonstrations recorded for each category with different parameters, focusing on RGB images and robot joint positions. This process was limited to cereals and tofu.

\begin{figure*}[htbp]
    \centering
    % First row
    \begin{subfigure}[b]{0.3\linewidth} % Adjust width to fit within a column
        \centering
        \includegraphics[width=\linewidth]{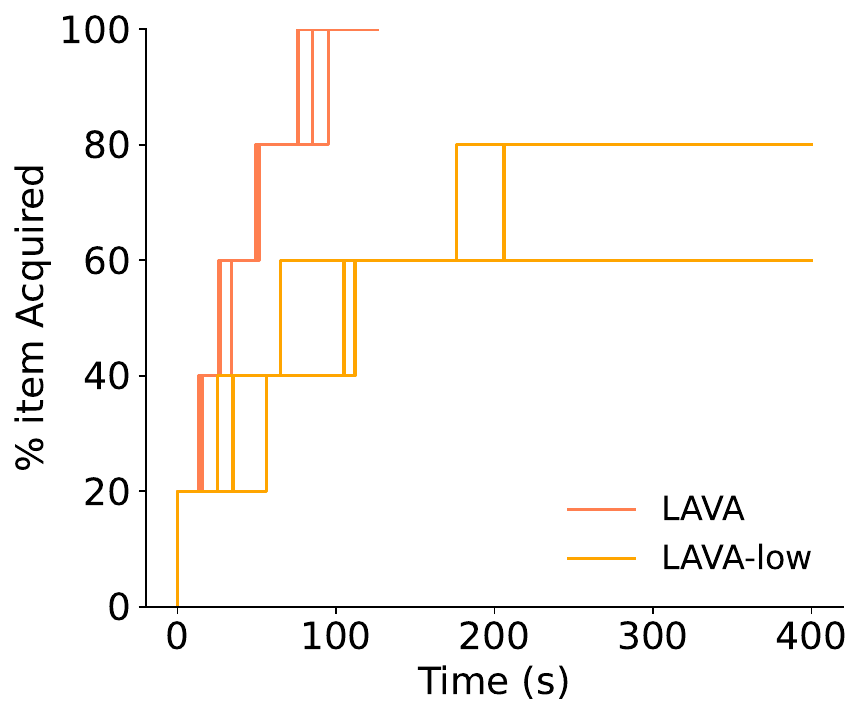}
        \caption{5 tofu chunks.}
        \label{fig:sub1}
    \end{subfigure}
    % \quad % Space out the subfigures slightly
    \begin{subfigure}[b]{0.3\linewidth} % Adjust width to fit within a column
        \centering
        \includegraphics[width=\linewidth]{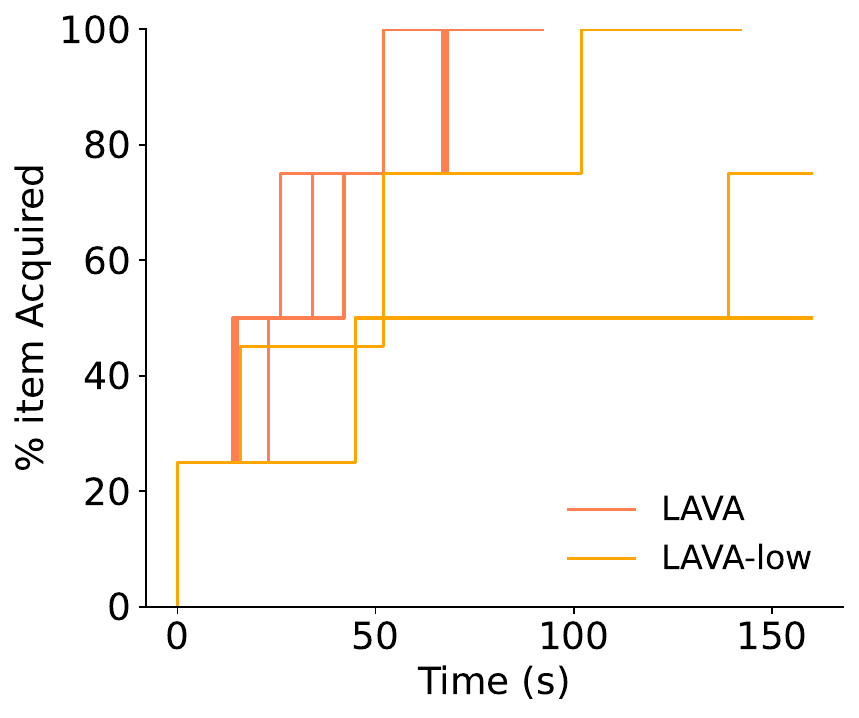}
        \caption{4 tofu chunks.}
        \label{fig:sub2}
    \end{subfigure}
    % \quad % Space out the subfigures slightly
    \begin{subfigure}[b]{0.3\linewidth} % Adjust width to fit within a column
        \centering
        \includegraphics[width=\linewidth]{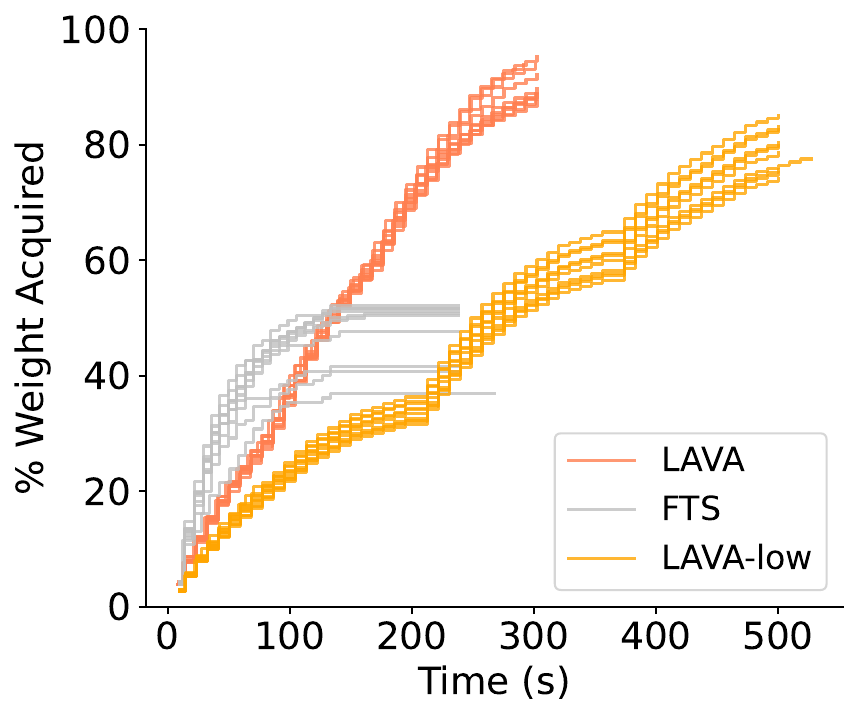}
        \caption{Cereals.}
        \label{fig:sub4}
    \end{subfigure}

    \caption{Individual trials comparison between \LAVA{}, LAVA-low and FTS. Subfigures (a) and (b) show the comparison with different tofu configurations, and (D) show the comparison with cereals.}
    % \vspace{-1em}
    \label{fig:subfigures}
\end{figure*}
% \vspace{-1.5em}
\subsection{Baselines}

In our study, we used two baselines, LAVA-low and Fixed Trajectory Scooping (FTS). For both baselines, the process begins with detecting the bowl in an RGB image using RetinaNet \cite{lin2017focal}. Upon identifying the bowl, we calculate its centroid and map this position to the robot's coordinate system. This allows the robot to move to the bowl's location, adjusting to a predetermined height and orientation. 

In the case of FTS, during tests with various food items in a stationary bowl position, wrist 2 of the robot arm is rotated by $-0.6$ radians to start the scooping action along a predefined trajectory.

Conversely, LAVA-low, employs the same low-level policy $\pi_L$ as \LAVA{} for scooping. For deformable food and fruit chunks, we stick with the wall-guided scooping trajectory for the R1 region(see Figure. \ref{fig:targetNet} for reference) and keep rotating the bowl every 45 degrees constantly so that the spoon can reach all the instances in the bowl near the wall and gets maximum coverage. In contrast for granular, liquid, and semi-solid foods we stick with direct scooping, adjusting its approach based on the depth of the food within the bowl. This adjustment occurs once a predefined depth threshold is reached, to effectively target the lower layers of food, ensuring thorough bowl clearance.

\subsection{Experimental Results}
In this section, we present and analyze the experimental results. We first present the success rate of LAVA's networks. Then, following the training of the hierarchical policy, we evaluate its performance on the robot and compare it with the baseline methods. We test across varied food items and varied food configurations, including granular food cereals, liquid food water, semi-solid yogurt, deformable tofu, and multi-medium soup with tofu chunks.
To assess performance, we employ the criteria of success rate, which indicates the successful scooping of food items from a bowl without spillage and breakage and successful long-horizon food acquisition by clearing the bowl efficiently. Instances where some spillage occurs are considered partial success.

\subsubsection{\LAVA{}'s Network Success rates}
ScoopNet achieved 100\% accuracy in choosing correct high-level primitives across 46 bowls, TargetNet accurately predicted bite targets at 87.9\% over 83 instances, and DepthNet successfully determined correct spoon depths for bite sizes at 85.7\% across 175 instances, demonstrating the \LAVA{} networks' effectiveness in robotic-assisted feeding.

%%%%%%%%%%%%%%%%%%%%%%%%%%%%%%%%%%%%%%%%%%%%%%%%%%%%%%%%%%%%%%%%%%%%%%%%%%%%%%%%%%%%%%
\subsubsection{Comparison with Baselines}
In our experimental analysis, we evaluate the success rates of \LAVA{} against two baseline models, Lava-low and FTS, across a variety of food types and scooping dynamics, as shown in Figures \ref{fig:subfigures_1} and \ref{fig:subfigures}. Our evaluation focused on several key metrics: efficiency (total time taken to clear the bowl), scoop size, and spillage for granular, semisolid, and liquid foods. For deformable foods and fruit chunks, we recorded configuration, number of scoop attempts, and instances of food breakage. In particular, for complex scenarios such as soup with tofu chunks, our assessment averaged efficiency, spillage, and breakage metrics.

\textbf{How do all the methods handle the challenge of scooping liquids, such as water and soup, which are prone to spillage?} The analysis, particularly visible in Figure \ref{fig:spillage}, reveals that both baseline models struggle with the fluidity of water and soup, leading to significant spillage. The FTS model, with its fixed end-effector orientation and height, is not equipped to adjust to the varying dynamics of liquid scooping, resulting in spillage and ineffective scooping. LAVA-low initially copes but struggles as water levels decrease, showing inefficiency in maintaining adequate scoop sizes. In contrast, \LAVA{} adeptly adjusts to real-time changes in food depth, achieving optimal scoop sizes and minimizing spillage for efficient bowl clearance.

\textbf{What about the acquisition of more solid, yet deformable food types, such as tofu?} Our findings, demonstrated in Figures \ref{fig:sub1}, \ref{fig:sub2}, and \ref{fig:breakage}, indicate that Both baselines encounter issues with deformable foods such as tofu, often resulting in food breakage. The FTS model’s rigid scooping motion damages the food, while Lava-low, despite managing to scoop, causes tofu to accumulate and break as shown by instances of food breakage in Figure \ref{fig:breakage} due to lack of strategic food prioritization based on subregions. \LAVA{}, however, prioritizes tofu chunks based on their subregion, aligning them for easier access and significantly reducing breakage, mimicking human scooping strategies.

 \textbf{How does each method fare in preventing spillage and ensuring efficient scoop attempts with solid foods such as fruit chunks?} The evaluation as visible in Figure \ref{fig:spillage} and \ref{fig:success_rate} reveals that the baselines are less adept with solid, irregularly shaped foods such as fruit chunks, prone to rolling or falling off the spoon. This issue is exacerbated for fruits with curved surfaces. \LAVA{}, employing an align-then-scoop strategy, ensures better alignment and significantly less spillage by adjusting to the fruit's shape for secure scooping.
 
 We see that \LAVA{} consistently outperforms the baselines, achieving higher success rates and more effective plate clearance. It surpasses FTS and Lava-low by adapting its strategy for efficient, minimal-breakage scooping across all tested food types, demonstrating the benefits of its hierarchical policy framework.  As expected, FTS and Lava-low, limited by their static approaches, fail to optimize for future scooping advantages, leading to increased breakage and inefficiency, especially without considering food prioritization and arrangement strategies.
 
\LAVA{}'s comprehensive strategy ensures efficient, adaptive, and precise food acquisition, significantly improving upon the limitations of existing models.

\subsubsection{Zero-shot Generalization}
\label{subsubsection:Zero-shot Generalization}
\begin{figure}[ht]
    \centering
    \includegraphics[width=0.5\textwidth]{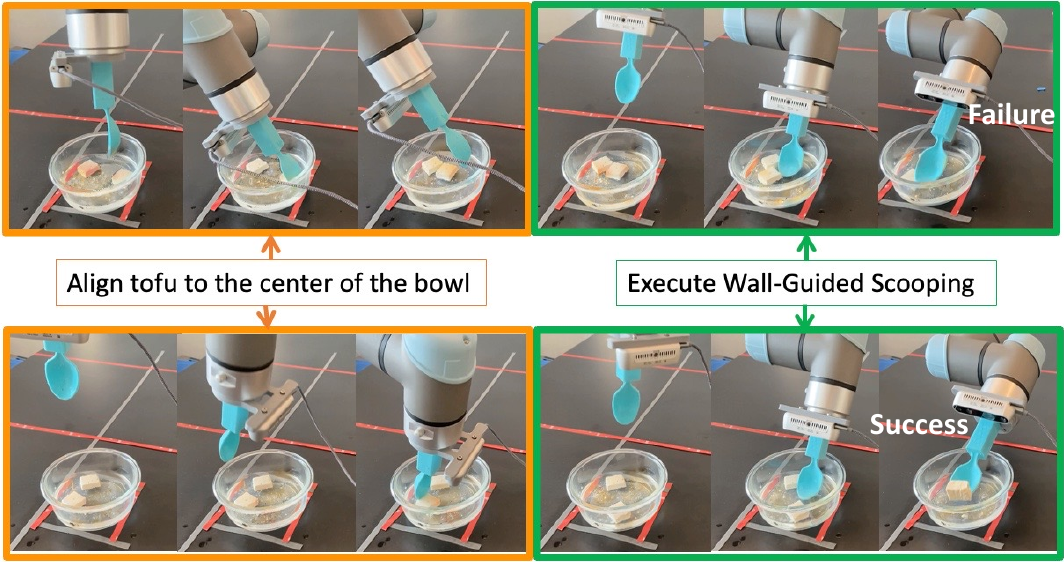}
    \caption{(Zero-shot) long-horizon food acquisition with tofu chunks in soup. The top sequence of images shows the spoon aligning the target tofu towards the bowl's center, which then drifts away during the scooping attempt due to the soup's fluidity. The bottom sequence shows the system's subsequent attempt to realign the tofu to the center, followed by a successful scooping action.}
    \label{fig:zero-shot}
\end{figure}

As detailed in Section \ref{subsection:data_collection}, our data collection process exclusively involved the transparent glass bowl containing granular cereals and tofu. However, we evaluated our approach to soup with tofu chunks and different food types such as liquid water and semi-solid yogurt, and solid apple chunks during testing. Remarkably, our approach demonstrates robust performance across these varied configurations, as depicted in Figure \ref{fig:subfigures_1} and \ref{fig:zero-shot}

Especially with soup and tofu chunks, scooping up both the solid pieces and the liquid at the same time is tricky. Our system, \LAVA{}, is designed to adjust to these challenges. Despite the tofu chunks' tendency to float away from the desired central scooping area, \LAVA{}'s adaptive strategy realigns and reorients to scoop the tofu effectively. Following the tofu acquisition, LAVA continues to adapt and clear the remaining soup, showcasing its capability to handle various food textures and types within the same meal, leading to efficient bowl clearance.

\section{Conclusion, limitation and Future work}
\label{section:conclusion}

In this work, we have developed and presented a hierarchical policy framework designed to enhance robotic systems' capability in the acquisition of diverse food types, ranging from liquids to solids and deformable items. Through integrating DepthNet, TargetNet, and ScoopNet, our approach leverages representation learning, alongside sophisticated planning and execution strategies, to address the challenges associated with the variability in food textures, sizes, and positions within the bowl.

Our experimental analysis demonstrates the framework's better performance in achieving better efficiency, minimum spillage, and breakage as well as adaptive food scooping compared to baseline models. Specifically, it showcases improvements in success rates across various food configurations. Despite the promising result towards generalization, limitations exist, particularly in handling thin, flat, or irregular foods needing specialized strategies. Future efforts will focus on broadening the action space for diverse food types and exploring efficient data acquisition methods, including leveraging internet video resources for complex food handling strategies in real-world scenarios.

\bibliographystyle{IEEEtran}
\bibliography{IEEEabrv,references}

\end{document}